%% file: Mawdoo3_DeepQ2Q_ArxiV.tex
\documentclass{article}

\usepackage{arabtex}
\usepackage{arxiv}

\usepackage[utf8]{inputenc} 
\usepackage[T1]{fontenc}    
\usepackage{hyperref}       
\usepackage{url}            
\usepackage{booktabs}       
\usepackage{amsfonts}       
\usepackage{nicefrac}       
\usepackage{microtype}      
\usepackage{lipsum}
\usepackage{cite}
\usepackage{mathtools}
\DeclarePairedDelimiterX{\inp}[2]{\langle}{\rangle}{#1, #2}
\usepackage{amsmath,amssymb,amsfonts}
\usepackage{algorithmic}
\usepackage{graphicx}
\usepackage{textcomp}
\usepackage{xcolor}
\usepackage{soul} 

\title{NSURL-2019 Shared Task 8: Semantic Question Similarity in Arabic}

\author{
  Haitham Seelawi 
  \\
  \texttt{haitham.seelawi@gmail.com} \\
   \And
 Ahmad Mustafa \\
  AI Department\\
  Mawdoo3 Ltd\\
  Amman, Jordan \\
  \texttt{ahmad.mustafa@mawdoo3.com} \\
   \AND
   Hesham Al-Bataineh \\
   AI Department\\
   Mawdoo3 Ltd \\
   Amman, Jordan \\
   \texttt{hisham.bataineh@mawdoo3.com} \\
   \And
   Wael Farhan\\
  AI Department\\
  Mawdoo3 Ltd\\
  Amman, Jordan \\
   \texttt{wael.farhan@mawdoo3.com} 
   \And
   Hussein T.~Al-Natsheh \\
   AI Department\\
   Mawdoo3 Ltd \\
   Amman, Jordan \\
   \texttt{h.natsheh@mawdoo3.com} \\
}

\begin{document}
\maketitle

\begin{abstract}
\input{sections/abstract.tex} 
\end{abstract}


\section{Introduction}
\label{sec:introduction}
\input{sections/introduction.tex}

\section{Related Work}
\label{sec:related}
\input{sections/related.tex}

\section{Dataset}
\label{sec:data}
\input{sections/data.tex}

\section{Participants and Systems}
\label{sec:models}
\input{sections/participants.tex}

\section{Results and Discussion}
\label{sec:setup}
\input{sections/results.tex}

\section{Conclusion}
\label{sec:conclusion}
\input{sections/conclusion.tex}

\section{Acknowledgement}
\input{sections/acknowledgement.tex}

\bibliographystyle{unsrt}  
\bibliography{references}  


\end{document}

%% file: sections/abstract.tex
Question semantic similarity (Q2Q) is a challenging task that is very useful in many NLP applications, such as detecting duplicate questions and question answering systems.
In this paper, we present the results and findings of the shared task (Semantic Question Similarity in Arabic).
The task was organized as part of the first workshop on NLP Solutions for Under Resourced Languages (NSURL 2019)
The goal of the task is to predict whether two questions are semantically similar or not, even if they are phrased differently.
A total of 9 teams participated in the task.
The datasets created for this task are made publicly available to support further research on Arabic Q2Q.

%% file: sections/introduction.tex





Semantic Textual Similarity (STS) is a core task in Natural Language Processing and Understanding (NLP/NLU). Simply put, STS is concerned with inferring the similarity in meaning between a pair of sentences. It should be mentioned that there are other levels of granularity for STS such as: Lexical (i.e. single words), full paragraphs or whole documents.


In this paper, we focus on the STS of a question pair (or \textit{Q2Q} Similarity).
We assume that if two questions have the same answers, then they are semantically similar. Otherwise, if the answers are different or partially different, then the pair is considered non-equivalent.

STS provides the basis for Question Answering systems (QA). As the name suggests, QA systems are computer systems which can answer questions posed in a natural language form. These questions can be of either factoid or non-factoid nature. Factoid questions can be defined as questions for which a complete answer can be given in 50 bytes or less (a few words)~\cite{DBLP:conf/naacl/SoricutB04}. These are typically questions that start with who, what, when or where, and have definitive answers. Non-factoid questions, on the other hand, require longer answers. They are mainly instructional or explanatory in nature.

One possible way to build QA systems using STS is having predefined questions along with their answers. When a user asks a question, a ranked list of these questions can be obtained, and based on that list, the best answer can be returned to the user. This method can be used, both, for factoid and non-factoid questions.

One important application to Q2Q is identifying duplicate questions in community question answering platforms (e.g., quora.com).
Users may ask questions that might be already asked and answered by the community.
Finding these duplicate questions saves the effort and time spent in answering already answered questions.
However, detecting duplicate questions is challenging because these questions, although are semantically similar, they might be phrased differently.
Moreover, dealing with the Arabic language in Q2Q similarity is challenging due to several factors. 
Arabic Q2Q datasets are scarce and limited in size.
Moreover, the Arabic language is one of the most morphologically rich languages.

In this paper, we present the results and findings of the shared task (Semantic Question Similarity in Arabic).
The task was organized as part of the first workshop on NLP Solutions for Under Resourced Languages (NSURL 2019)\footnote{\url{http://nsurl.org/}}
The goal of the task is to predict whether two questions are similar or not.
A total of 9 teams participated in the task.
Among them, 4 teams have provided description papers, which are included in the NSURL workshop proceedings.

The rest of this paper is organized as the following. 
In Section~\ref{sec:related}, we discuss previously published work relating to Q2Q in Arabic.
Section \ref{sec:data} provides an overview of the datasets used in the task.
Next, in Section \ref{sec:participants} we briefly describe the participants and the approaches they propose. 
Then we discuss the experiments and analyze the results of the competition in Section \ref{sec:results}. 
Finally, we conclude in Section \ref{sec:conclusion}.

%% file: sections/related.tex
Despite its importance and utility in NLP applications, research on STS at the level of sentences and higher, has only picked up steam in the past ten years~\cite{DBLP:journals/corr/abs-1708-00055}. Nonetheless a lot has been accomplished since, but mainly in the English language. In the case of Arabic, there is plenty of room for new research to advance the current state of the art in this regard~\cite{alianarabic}. Therefore, most of our review below will focus on methods developed and used in English mainly, which might not be directly applicable to Arabic.

Some of the earliest methods used in the field made extensive use of so-called knowledge-based semantic similarities between words~\cite{DBLP:journals/cys/MajumderPGP16}. These can be thought of as lexical databases that model the semantic relationships of different words, taking into consideration their different senses. At the center of these databases is the concept of “synsets”, which are groups of words that refer to a specific concept. The most popular such database is WordNet~\cite{DBLP:journals/lre/MillerF07}. With the assistance of word alignment methods, various meaningful numerical features pertaining to the lexical units comprising a pair of sentences can be obtained from WordNet. Combined with other textual features, such as Part of Speech (POS), and Term Frequency - Inverse Document Frequency (TF-IDF), and fed into strong classifiers, such methods obtain very good results, albeit on closed domains of assessment~\cite{DBLP:conf/semeval/SaricGKSB12,DBLP:journals/bioinformatics/SoganciogluOO17,DBLP:conf/acl/PilehvarJN13}. Nonetheless, it can be easily seen that the construction of such databases, is very expensive in terms of human effort.

Semantic relationships can be modeled using another class of methods named Word Vector Representations (WVR). One of the biggest advantages of such methods is that they are typically trained in an unsupervised manner, making their construction very cheap in terms of human annotation. Some of these methods include Word2Vec \cite{DBLP:conf/nips/MikolovSCCD13}, Glove \cite{pennington2014glove}, ELMo \cite{peters2018deep} and BERT \cite{devlin2018bert}. These word representations significantly boost the performance of machine learning algorithms~\cite{DBLP:conf/nips/MikolovSCCD13}, especially deep learning-based approaches.

One of the earlier and more basic methods of using WVR in STS, consisted in pooling the corresponding dimensions of tokens in a given sentence, using a specific pooling method, such as the average, or the maximum, to obtain a sentence level representation from WVR. The representation of each sentence in the pair would then serve as the input into a classifier or a predefined measure of similarity. One of the obvious advantages of such a method is its simplicity, and that it can be readily used in many classes of machine learning algorithms. However, it is apparent that by using pooling, we are losing all the information about the order of tokens in the original sentences, which definitely matters in defining the meaning of a sentence. Additionally, by using pooling methods, we are assuming that words and sentences can be represented using the same space size, which is a limitation of such a method~\cite{DBLP:journals/corr/abs-1901-10444}.

One relatively recent advancement in STS, which accounts for the shortcomings of the pooling methods is the Siamese Recurrent Architecture~\cite{DBLP:conf/aaai/MuellerT16}. By using two Recursive Neural Networks (RNNs), with shared weights, the pair of sentences are encoded into a higher dimensional space than the WVR used for the constituent tokens. Given the sequential nature of RNNs, this encoding takes into account the order of tokens in each sentence. The encoding is then fed into a feedforward dense neural network, with a value between 0 and 5 to predict the semantic similarity of the pair. One of the advantages of this method when it comes to inference, is that it can be used to produce a sentence level representation, which, with the use of a simple distance matrices, can be used to measure the similarity between two sentences without the need for the feedforward step~\cite{DBLP:conf/rep4nlp/NeculoiuVR16}. This translates to much higher scalability in industrial applications. Another advantage is that it can be modified to account for errors in spelling~\cite{DBLP:conf/rep4nlp/NeculoiuVR16}. Nonetheless, a major drawback of this method is that it requires a substantial amount of annotated data for training.

One method which overcome this limitation is Skip-thought Vectors (SV)~\cite{DBLP:conf/nips/KirosZSZUTF15}, which learn to embed text at the level of sentences, by training on continuous text (e.g. books and articles) in an unsupervised fashion. The representations can then used as feature inputs with the method of choice to predict the STS score. However, training SV requires very long period of time (it took about one month back in 2015~\cite{DBLP:journals/corr/abs-1901-10444}).

One problem that most sequential deep learning methods suffer from is that the longer the sequence of text to encode is, the less efficient the representation becomes~\cite{olah2016attention}. This problem has been recently tackled by exploiting the attention mechanism in deep learning architectures. With the use of multi-head attention mechanism in constructing sentence embeddings, the state of the art of NLP in many STS dependent tasks has been significantly increased~\cite{DBLP:journals/corr/LinFSYXZB17}.

Another recent and novel development pertaining to STS, makes use of conversational data~\cite{DBLP:conf/rep4nlp/YangYCKCPGSSK18}. The premise here is that sentences that are semantically related, will elicit similar responses in a conversation. However, an obvious shortcoming of such a method is that it is by design geared toward conversational tasks, as opposed to tasks that are factual by nature.

As it stands now, the state of the art in STS are Universal Sentence Encoders (USE)~\cite{DBLP:journals/corr/abs-1803-11175}. These encoders are trained on a wide variety of data types and tasks (i.e. using different signals such as entailment and SV like signals), with the idea of transfer learning at their heart. Under the hood, USEs can be powered by one of two deep learning architectures; the first is a transformer network, while the other is a deep averaging network. The main difference between these two versions, is that with the former, higher accuracies can be achieved, but with longer training times, whereas for the latter, training is less computationally intensive, at the expense of some accuracy in the final outcome.

%% file: sections/data.tex
Despite the fact that there is a number of public datasets for QA in English language (such as SQuAD~\cite{DBLP:journals/corr/RajpurkarZLL16} and CoQA~\cite{DBLP:journals/corr/abs-1808-07042} to name a few, there is a dearth of such datasets in Arabic.
Therefore, we have developed a dataset\footnote{\url{https://ai.mawdoo3.com/nsurl-2019-task8}} of $15,712$ pairs of questions, that were annotated and verified by an internal team of qualified natural language annotators. Each pair has a ground truth of either “0” (no semantic similarity), or “1” (denoting semantically similar pairs).
We have randomly selected 11,997 pairs for training and used the remaining 3,715 for testing. We made sure that the collected data is balanced, where the number of similar question pairs is comparable with the not similar ones. Table \ref{tab:dataset} shows a detailed statistics of Mawdoo3 Q2Q dataset.

\begin{table} [tb]
\centering
\caption{Mawdoo3 Q2Q dataset statistics.}
\label{tab:dataset}
\begin{tabular}{|c|c|c|c|}
\hline
\textbf{Set} & \textbf{Similar} & \textbf{Not Similar} & \textbf{total} \\
\hline
Train & 5,397 & 6,600 & 11,997 \\
\hline
Test & 1,718 & 1,997 & 3,715 \\
\hline
Total & 7,115 & 8,597 & 15,712 \\
\hline
\end{tabular}
\end{table}

These questions were designed specifically to contain a balanced number of factoid and non-factoid questions. Additionally, great care was taken in assuring that the pairs of questions have varying STS and LS similarity, in a way that mimics the population of questions asked on the internet by Arabic language users. For example:
\begin{center}
    \begin{RLtext}
        mn hw r'Is AlwlAyAt Almt.hdT Ala'mrykyT?
    \end{RLtext}
\end{center}
which translates to ``Who is the president of the United States of America?''.

\begin{figure}[tb]
\label{fig:length_histogram}
\centering
\includegraphics[scale=0.50]{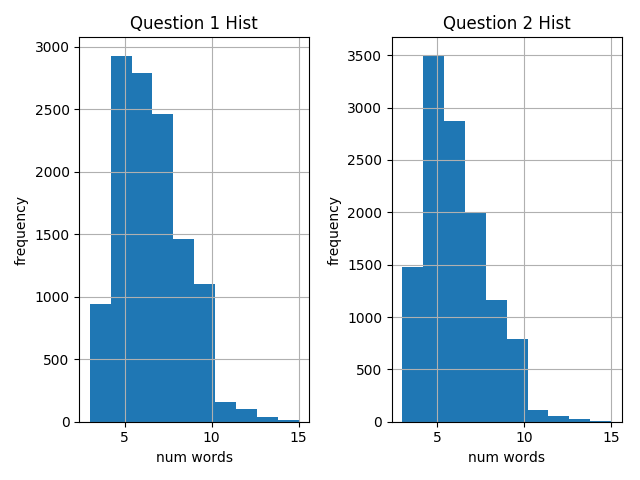}
\caption{Distribution of question lengths (word count) in Mawdoo3 Q2Q dataset. The figure on the left shows Question 1 histogram, and Question 2 on the right.}
\end{figure}

Table \ref{tab:dataset} lists a small sample of the dataset. The dataset consists of 3 fields, i.e. \textit{question1} containing the text for one of the question pairs, \textit{question2} containing the text of the second question, and \textit{label} which is either 1 if question1 and question2 have a similar answer, or 0 if their answers are different. Figure \ref{fig:length_histogram} shows a histogram for a number of words per question against frequency. It can be seen that the maximum question length is 15 words and that the distribution of both \textit{question1} and \textit{question2} is almost the same.

\begin{table*} [tb]
\centering
\caption{Sample of the Mawdoo3 Q2Q dataset. The dataset is composed of three columns. The first two are text fields containing question1 and question2 while the third column shows the label.}
\label{tab:sample}
\begin{tabular}{|c|c|c|}
\hline
\textbf{question1} & \textbf{question2} & \textbf{label} \\
\hline
\RL{mA hy Al.trq Al.s.hy.hT llA`tnA' bAl.hAml?} & \RL{kyf Ahtm b.tfly?} & 0 \\
\hline
\RL{mA .tryqT t.h.dyr m.h^sy AlkwsA ?} & \RL{mn .trq t.h.dyr m.h^sy AlkwsA?} & 1 \\
\hline
\RL{fy 'y `Am wld twfyq Al.hkym?} & \RL{'yn wld twfyq Al.hkym?} & 0 \\
\hline
\RL{mA .tryqT t.h.dyr AAlmhlbyxxT bjwz Alhnd?} & \RL{kyf A.h.dr AlmhlbyxxT bjwz Alhnd?} & 1 \\
\hline
\RL{mA .tryqT t.h.dyr Alkyk Alm.h^sy bAlkrymT ?} & \RL{mn .trq t.h.dyr AlkrymT?} & 0 \\
\hline
\RL{mA hy .h.swAt AlmrArT?} & \RL{mA hy .h.s_A AlmrArT?} & 1 \\
\hline
\RL{kyf A.h.drAlm.sAbyb m` Almk^sn?} & \RL{mn .trq t.h.dyr Alm.sAbyb  Alm.h^sy?} & 0 \\
\hline
\RL{mA hw Almwt?} & \RL{mA 'jml mA qyl bAlmwt?} & 0 \\
\hline
\RL{fy 'y `Am buny brj _hlyfT?} & \RL{'yn ywjd brj _hlyfT?} & 0 \\
\hline
\RL{mA .tryqT t.h.dyr `jynT AlbytzA bAl.hlyb ?} & \RL{mn .trq t.h.dyr `jynT AlbytzA  ?} & 0 \\
\hline
\RL{mA m`n_A AljhAd?} & \RL{mA 'nwA` AljhAd?} & 0 \\
\hline
\RL{lmA_dA mydAn bykAdyly yj_db Alk_tyr mn AlsyA.h?} & \RL{mA Asm 'hm m`lm syA.hy fy bry.tAnyA?} & 0 \\
\hline
\RL{km ybl.g .twl tm_tAl Almsy.h AlfAdy?} & \RL{mA hw .twl Altm_tAl AlfAdy?} & 1 \\
\hline
\RL{'il_A km y.sl ArtfA` 'bw Alhwl Almwjwd fy m.sr?} & \RL{km ybl.g `dd skAn m.sr?} & 0 \\
\hline
\RL{mn hw Almdyr Al`Am ?} & \RL{mA hw t`ryf Almdyr Al`Am ?} & 1 \\
\hline
\RL{mA hy AdArT AlA`mAl ?} & \RL{mA hy mjAlAt AdArT AlA`mAl ?} & 0 \\
\hline
\RL{mAhw Alkwlystrwl?} & \RL{mA t`ryf Alkwlystrwl?} & 1 \\
\hline
\RL{mA hy 'hmyT AlAst_tmAr ?} & \RL{Al_A mA_dA yhdf AlAst_tmAr ?} & 1 \\
\hline
\end{tabular}
\end{table*}

%% file: sections/participants.tex
The shared task was managed using a Kaggle competition platform\footnote{\url{https://www.kaggle.com/c/nsurl-2019-task8}} for registration and results submissions. We have published a baseline\footnote{\url{https://github.com/mawdoo3/q2q\_workshop}} that the participants can reproduce on the same dataset.

A total of 9 teams participated in this task, with total submissions of 547, and an average of more than 60 submissions per team. In this section, we report the methodologies used for four different teams.

\subsection{The Inception}
\input{sections/systems/inception.tex}

\subsection{Tha3aroon}
\input{sections/systems/tha3aroon.tex}

\subsection{onekaggler}
\input{sections/systems/onekaggler.tex}

\subsection{Speech Translation}
\input{sections/systems/speechtranslation.tex}

%% file: sections/systems/inception.tex
The Inception team members applied different deep learning approaches, including BERT model \cite{devlin2018bert}. They fine-tuned the multilingual BERT model \cite{devlin2018bert} on the sentence similarity task.

They tried various combinations of hyperparameters. For the set of parameters that made the best predictions, they repeated the experiment with different random seeds, then created an ensemble model by voting between the prediction results of these experiments. The ensemble that is composed of 3 models performed better on the public dataset while 4, 5, and 6 models have better scores on the private dataset.

%% file: sections/systems/tha3aroon.tex
Tha3aroon team did heavy work on the dataset level before building the model. First, they made sure that punctuation marks are separated from the words by making sure that characters surrounding the punctuation marks are spaces. Next, they augmented the dataset 4 different methods:

\begin{itemize}
  \item \textbf{Positive Transitive:} If A is similar to B, and B is similar to C, then A is similar to C.
  \item \textbf{Negative Transitive:} If A is similar to B, and B is NOT similar to C, then A is NOT similar to C. This rule combined with the previous one generates 5,490 extra examples (17,487 total).
  \item \textbf{Symmetric:} If A is similar to B then B is similar to A, and if A is not similar to B then B is not similar to A. This rule doubles the number of examples to 34,974 in total.
  \item \textbf{Reflexive:} By definition, a question A is similar to itself. This rule generates 10,540 extra positive examples (45,514 total) which help balance the positive and negative examples.
\end{itemize}

After the augmentation process, the training data contains 45,514 examples (23,082
positive examples and 22,432 negative ones).

To build meaningful representations for the input sequences, they used Arabic ELMo \cite{peters2018deep} pre-trained model \footnote{https://github.com/HIT-SCIR/ELMoForManyLangs} to extract contextual words embeddings and feed them as an input to the model. The model then consists of three components:

\begin{enumerate}
    \item \textbf{Sequence representation extractor:} which takes the ELMo embeddings related to each word in the question as an input and feeds them to two special kinds of LSTM layers called Ordered Neurons LSTM (ON-LSTM) \cite{shen2018ordered} and applies sequence weighted attention \cite{felbo2017using} on the outputs of the second ON-LSTM layer to get the final question representation, this component uses the same weights to compute representations for pair questions.
    \item \textbf{Merging layer:} After extracting the representations related to each question, they merged the representations using a pairwise squared distance function applied on the pair questions representation vectors. 
    \item \textbf{Deep neural network:} Consisting of four fully-connected layers that take the merged representation vector as an input and predicts the label using a sigmoid function as an output.
\end{enumerate}

%% file: sections/systems/onekaggler.tex
\begin{figure}[tb]
\label{fig:onekaggler}
\centering
\includegraphics[scale=0.35]{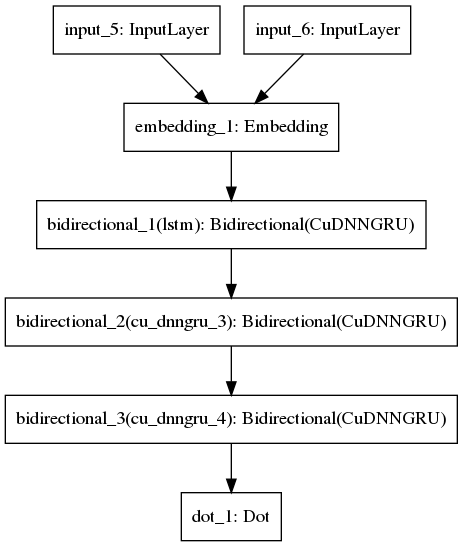}
\caption{onekaggler model}
\end{figure}

The onekaggler team has built a neural network model illustrated in Figure \ref{fig:onekaggler}. The model consists of two input layers for question1 and question2,
a shared trainable word embedding layer, using Word2Vec model \cite{DBLP:conf/nips/MikolovSCCD13}, initialized with Aravec tweets\_cbow\_300 embedding model~\cite{aravec},
and a stack of 3 bidirectional GRU layers with 256, 128, 64 hidden nodes, respectively.
The output layer is the dot product (which calculates cosine similarity) between the outputs of the last layer of question1 and question2. 
The team uses mean-squared-error as a loss function alongside with Nesterov Adam optimizer.
They achieve 99\% accuracy on the validation set and under 94\% on the test set. 

%% file: sections/systems/speechtranslation.tex
The Speech Translation team members have gathered feature set using sklearn's Vectorizer Analyzer with three setups; \texttt{word-level}, \texttt{char-level}, and \texttt{char\_Wb-level}.
They have examined the use of n-grams (1, 2, 3, 4, and 5) for the three setups.
As a preprocessing step, they applied punctuation removal, stop words filter, and text normalization.
These features, combined with word stemming and POS tagging, are used for model training and testing.
The team has compared the performance of a set of classifiers: BNB, LogReg, LSVM, MNB, PassAgg, PRP and SGD as well as CNN. 
The best performance is achieved by LSVM classifier.

%% file: sections/results.tex
Table \ref{tab:results} shows a summary of results for the participating teams. 
The Inception team has topped the list by achieving an accuracy score of 0.9592 using BERT models.
ELMo model built by Tha3aroon scored second with an accuracy of 0.9485. This model was trained using the augmented dataset of 45,514 data samples.
onekaggler team has scored third among all participants with 0.9481 accuracy using a stack of three Bidirectional GRUs.
Speech Translation team has used 1 to 5 n-grams of words and characters and has experimented with several classifiers to score 0.8270, achieving the 7$^{th}$.

\begin{table} [htb]
\centering
\caption{Results for Semantic Question Similarity in Arabic. The table shows the 9 teams who participated in the workshop sorted in descending accuracy score.}
\label{tab:results}
\begin{tabular}{|c|c|c|}
\hline
\textbf{\#} & \textbf{Team Name} & \textbf{Score} \\
\hline
1 & The Inception & 0.95924 \\
\hline
2 & Tha3aroon & 0.94848 \\
\hline
3 & onekaggler & 0.94809 \\
\hline
4 & Ayat Abedalla & 0.91311 \\
\hline
5 & Dan Ofer & 0.89465 \\
\hline
6 & heza & 0.85736 \\
\hline
7 & Speech Translation & 0.82698 \\
\hline
8 & AtyNegar & 0.82583 \\
\hline
9 & Eyad Sibai & 0.71434 \\
\hline
\end{tabular}
\end{table}

One of the main takeaways is that BERT model accuracy is higher than ELMo model even when it was fine-tuned on an augmented dataset. The BERT model learns the representation of subwords while ELMo is character based model that uses convolution layers to learn word embeddings that handle out of vocabulary words. The reported results show that BERT is able to strike a good balance between a character based and word based representations and capture the word semantics for the problem of Arabic Q2Q.

Both of ELMo and BERT were able to outperform the traditional Word2Vec embeddings that is not able to capture contextual semantics nor learns subword embeddings. This proves that Arabic language (a morphologically rich language) complicates the training phase for such models because it needs to learn a completely new embedding for each morphology and is unable to generalize learnings across word variations. A word root in the Arabic language can have up to 1000 variation, Word2Vec needs to learn a number of weights equal to the number of variations multiplied by the vector size, while BERT and ELMo will only need to learn the word prefixes, roots, and word prefixes.

An interesting experiment would be to train BERT on the augmented data developed by Tha3aroon.

%% file: sections/conclusion.tex
In this paper, we described the Arabic question similarity (Q2Q) shared the task that was organized in the workshop on NLP Solutions for Under Resourced Languages (NSURL 2019). The dataset of the shared task was made publicly available as a benchmark of this NLP task. 
A total of 9 teams participated in the task in which we provided a brief description of 4 of them who submitted their system description. 
The use of recent approaches in text embedding, i.e., BERT and ElMo, was a big factor in obtaining the best performing results.
Another approach was using data augmentation that boosted up the performance. 
Also, an approach of using a neural network with Adam optimizer and an input layer that is initialized with pre-trained word vectors of the question pair was a well-performing solution.
The ample number of participants in this workshop is an indication of the importance and interest in the Arabic language and Arabic semantic textual similarity. As future work, we would like to consider extending the task to news headlines as well as article titles.

%% file: sections/acknowledgement.tex
We would like to thank Mawdoo3 AI data annotation team members who contributed to build and release Mawdoo3 Q2Q Dataset: Riham Badawi, Lana AlZaatreh, Raed AlRfouh, and Dana Barouqa. We would also like to thank Mawdoo3\footnote{\url{ai.mawdoo3.com}} for making the datasets created for this task publicly available to support further research on Arabic Q2Q.